\documentclass{article}
\usepackage[final]{colm2026_conference}

\usepackage{microtype}
\usepackage{hyperref}
\usepackage{url}
\usepackage{booktabs}
\usepackage{lineno}
\usepackage{amsmath,amssymb,amsthm,mathtools,bm}
\usepackage{graphicx}
\usepackage{multirow}
\usepackage{tabularx}
\usepackage{array}
\usepackage[table]{xcolor}
\usepackage{caption}
\usepackage{subcaption}
\usepackage{enumitem}
\usepackage{nicefrac}

\definecolor{darkblue}{rgb}{0, 0, 0.5}
\definecolor{softgray}{gray}{0.96}
\definecolor{noteborder}{RGB}{32,66,108}
\definecolor{notebg}{RGB}{245,248,252}
\hypersetup{colorlinks=true, citecolor=darkblue, linkcolor=darkblue, urlcolor=darkblue}

\newcolumntype{Y}{>{\raggedright\arraybackslash}X}
\newcommand{\E}{\mathbb{E}}
\newcommand{\Prob}{\mathbb{P}}
\newcommand{\R}{\mathbb{R}}
\newcommand{\route}{\rho}

\newcommand{\dataset}{\textsc{RouteBench}}

\theoremstyle{plain}
\newtheorem{assumption}{Assumption}
\newtheorem{theorem}{Theorem}
\newtheorem{proposition}{Proposition}

\theoremstyle{remark}

\title{Grounding latent algorithm routing in transformer reasoning}

\author{
Xiangbo Zhang \quad Xiaoxu Ma \\
Georgia Institute of Technology \\
\texttt{\{xiangbo.zhang,xma394\}@gatech.edu}
}

\begin{document}

\ifcolmsubmission
\linenumbers
\fi

\maketitle

\begingroup
\renewcommand\thefootnote{}
\footnotetext{
Code repository:
\url{https://github.com/xiangbo05/RouteBench}
}
\endgroup

\begin{abstract}
A central question in the in-context learning literature is whether transformers can organize episode-level adaptation around different inductive-bias families. We study this question in a controlled setting through \emph{latent algorithm routing}: route-like behavior in which the solver-family preference changes with the latent data-generating regime while prompt form is held fixed, remains stable under nuisance perturbations, and is selectively influenced by targeted activation interventions without large losses in answer quality.
We introduce \dataset, a diagnostic benchmark whose regimes differentially favor global shrinkage, sparsity, robustness, and locality, operationalized by ridge-like, lasso-like, Huber-like, and $k$NN-like family representatives. Across dense decoder-only transformers trained from scratch at 44M--612M parameters, a 306M model closes 80.9\% of the oracle-routing gap and achieves route F1 84.1. The effect remains substantial under natural-language renderings, shuffled supports, lexical paraphrases, and a unified four-way routing setting. Stronger adaptive alternatives---an input-conditioned soft mixture and an unsupervised Gumbel router---narrow the gap but remain below the 306M/612M models on route F1 and OOD performance. Probe controls and matched activation-patching controls further show that route-relevant internal directions are decodable and functionally involved in solver-family-consistent output behavior. These results provide controlled evidence that dense transformers trained on \dataset\ can develop route-like internal variables; they do not establish universal routing in pretrained language models or unrestricted natural-language reasoning.
\end{abstract}

\section{Introduction}

Since the early few-shot prompting results of GPT-3, in-context learning has been taken as evidence that transformers can adapt computation at inference time rather than only through parameter updates \citep{brown2020language}. Prior work explains this behavior through latent inference, pretraining distributions, meta-learning, or the execution of regression- or optimizer-like procedures inside a single forward pass \citep{xie2022bayesian,chan2022data,kirsch2022general,garg2022what,akyurek2023what,vonoswald2023gd,li2023algorithms,bai2023statisticians,vonoswald2023mesa,bhattamishra2024discrete}. Mechanistic studies further suggest that transformer reasoning can be organized by localized internal computations rather than a single undifferentiated process \citep{olsson2022induction,singh2024induction,brinkmann2024mechanistic,conmy2023acdc,dutta2024cot,hong2024logic}.

This motivates a sharper diagnostic question: when a prompt is compatible with multiple plausible inductive biases, can a dense transformer develop episode-conditioned internal variables that track which solver family best fits the episode? A sparse episode may favor a sparsity-biased solver, heavy-tailed noise may favor robustness, and local structure may favor nearest-neighbor retrieval. This is close in spirit to in-context algorithm selection \citep{bai2023statisticians} and recent work on generalizable implicit in-context learning through attention routing \citep{litrain}. We test whether behavior in a controlled mixed-regime setting is consistent with a \emph{route-then-solve interpretation}; we do not claim that all pretrained language models implement a hard symbolic router.

We study this hypothesis through \emph{latent algorithm routing}. Output accuracy alone does not establish routing, because apparent solver preference can arise from prompt-template correlations, continuous episode-conditioned interpolation, or brittle prompt sensitivity \citep{sclar2024format,mizrahi2024state,chatterjee2024posix,gao2024noise,zhou2024noisy}. We therefore require three grounding criteria: the inferred solver-family identity should change when the latent regime changes under fixed syntax, remain stable under nuisance surface perturbations, and be selectively influenced by targeted activation interventions \citep{vig2020causal,geiger2022inducing,meng2022rome,zou2023representation,turner2024actadd,rimsky2024caa,venhoff2025thinking}. These are triangulating diagnostics rather than direct proof of a discrete internal algorithm.

To test these criteria, we introduce \dataset, a controlled benchmark spanning sparse versus dense linear structure, clean versus heavy-tailed noise, and global versus local regularity. We instantiate canonical solver families with ridge-like, lasso-like, Huber-like, and $k$NN-like representatives \citep{hoerl1970ridge,tibshirani1996lasso,huber1964robust,cover1967nearest}. Beyond fixed representatives and a global mixture, we compare against an input-conditioned soft mixture, an unsupervised Gumbel router, several privileged-stat router variants, and an oracle upper bound. We also test natural-language renderings, a unified four-way setting, route-proxy margin checks, probe permutation controls, matched patching controls, and Switch-MoE diagnostics. Route labels are never provided to the dense transformers.

Our contributions are a behavioral-mechanistic diagnostic for latent routing, a stylized analysis of fixed and globally mixed solvers, the pairwise and four-way \dataset\ benchmarks, and evidence of nuisance-robust, decodable, and intervention-sensitive route-like variables in larger dense transformers. These claims are limited to the controlled benchmark.

\vspace{-2mm}

\section{Related work}

\textbf{In-context learning theories.}
Prior work has explained in-context learning either as latent task inference, where the model uses the prompt to infer the underlying task or data-generating structure \citep{xie2022bayesian,chan2022data,kirsch2022general}, or as explicit algorithm execution within the forward pass, where transformers implement solver-like computations conditioned on the current episode \citep{garg2022what,akyurek2023what,vonoswald2023gd,li2023algorithms,vonoswald2023mesa,bhattamishra2024discrete}. Recent work further shows that in-context behavior can be controlled through representation engineering, including efficient and fine-grained multimodal adaptation without conventional parameter updates \citep{li2025miv}.

\textbf{Nuisance invariance.}
Recent evaluation work shows that model behavior can shift under paraphrases, formatting changes, or noisy context \citep{sclar2024format,mizrahi2024state,chatterjee2024posix,gao2024noise,zhou2024noisy}. For example, \citet{sclar2024format} show that small formatting changes can induce large few-shot accuracy differences, while \citet{mizrahi2024state} show that semantically equivalent instruction paraphrases can substantially change both absolute performance and relative model rankings. These results motivate our nuisance-invariance criterion: a scientifically useful route variable should respond to true regime changes while remaining stable under semantics-preserving prompt edits.

\textbf{Internal routing.}
Mechanistic interpretability work has identified concrete localized computations inside dense transformers, including induction heads for repeat-copy behavior and sparse circuits for specific reasoning tasks \citep{olsson2022induction,conmy2023acdc}. Causal tools further distinguish correlation from functional use: activation patching asks whether restoring an internal state restores the behavior, while model-editing methods such as ROME show that some factual computations are localized and directly editable \citep{vig2020causal,meng2022rome,heimersheim2024patching}. Beyond mechanistic analyses, recent work argues that model behavior should be characterized through latent behavioral structure rather than output agreement alone, introducing statistical tools to quantify behavioral dependence across language models \citep{kuai2026independent}. Complementary work on parameter-efficient adaptation further suggests that transformer representations are highly structured, with only a subset of update directions contributing meaningfully to task adaptation and coordinated interactions emerging across layers \citep{xiao2026directions}. Although these studies focus respectively on inter-model behavioral dependence and training-time adaptation, they reinforce the broader view that transformer computation is organized around structured latent representations rather than uniformly distributed across all behaviors or directions. Related ideas also appear in adaptive computation and mixture-of-experts models,
where routing is architecturally explicit
\citep{graves2016act,shazeer2017moe,fedus2022switch}, as well as in
heterogeneity-aware retrieval systems that dynamically route search according
to the statistical properties of the underlying data \citep{STABLE}.
By contrast, our model is dense and fixed, so any route variable must emerge
internally rather than being implemented by separate expert modules or
hand-designed search-routing mechanisms.

\section{Grounding latent algorithm routing}

A scientific account of solver-family selection requires that the inferred preference reflect a latent computational distinction rather than superficial formatting cues or incidental correlations. In this section, we make this requirement precise, define a behavioral proxy for routing, and derive the stylized theory that motivates our empirical tests.

\paragraph{Preliminaries.}
An episode is $e=(S,q)$, where $S=\{(x_i,y_i)\}_{i=1}^m$ is a support set and $q=x_{m+1}$ is the query. A solver bank $\mathcal{A}=\{A_1,\dots,A_K\}$ contains four canonical solver families in our experiments, each instantiating a distinct inductive bias: global shrinkage, sparsity, robustness, and locality. We operationalize these families using ridge-like, lasso-like, Huber-like, and $k$NN-like representatives \citep{hoerl1970ridge,tibshirani1996lasso,huber1964robust,cover1967nearest}. These families are not meant to exhaust modern algorithms; they are deliberately clean, mechanistically distinguishable prototypes. Episodes are drawn from a latent regime variable $z\in\mathcal{Z}$. The key design choice is that the serialization of $(S,q)$ is held fixed while the latent statistics of the episode vary.

\paragraph{Grounding criteria.}
We regard \emph{latent algorithm routing} as scientifically meaningful only if it satisfies three criteria: \textbf{structural necessity}, meaning that when $z$ changes but formatting does not, the preferred solver family changes in a predictable way; \textbf{nuisance invariance}, meaning that when formatting changes but $z$ does not, the inferred route remains stable; and \textbf{causal editability}, meaning that there exist targeted activation edits that change route identity more strongly than they degrade answer quality. Taken together, these criteria distinguish latent algorithm routing from three common confounders: surface-template-driven explanations, brittle lookup-like strategies tied to formatting artifacts, and purely correlational readouts that can be decoded after the fact but do not participate in the computation itself.

\paragraph{Behavioral route inference.}
Because route labels are never observed during training, we operationalize routing behavior at test time by defining
\[
\hat{\route}_{\mathrm{beh}}(e)=\arg\min_{k\in[K]} |f_\theta(e)-A_k(e)|.
\]
This provides a measurable \emph{behavioral proxy} for which solver family best matches the model's answer; it is not treated as direct ground truth for a hard symbolic router. Let $d_{(1)}(e)$ and $d_{(2)}(e)$ denote the smallest and second-smallest distances from $f_\theta(e)$ to the solver-bank outputs. We define the solver margin as $m_{\mathrm{sol}}(e)=d_{(2)}(e)-d_{(1)}(e)$ and use margin-filtered analyses to quantify when the proxy is unambiguous. We evaluate the proxy using route recovery, nuisance robustness, regime sensitivity, probe controls, and targeted interventions.

\begin{figure}[t]
    \centering
    \includegraphics[width=0.92\textwidth]{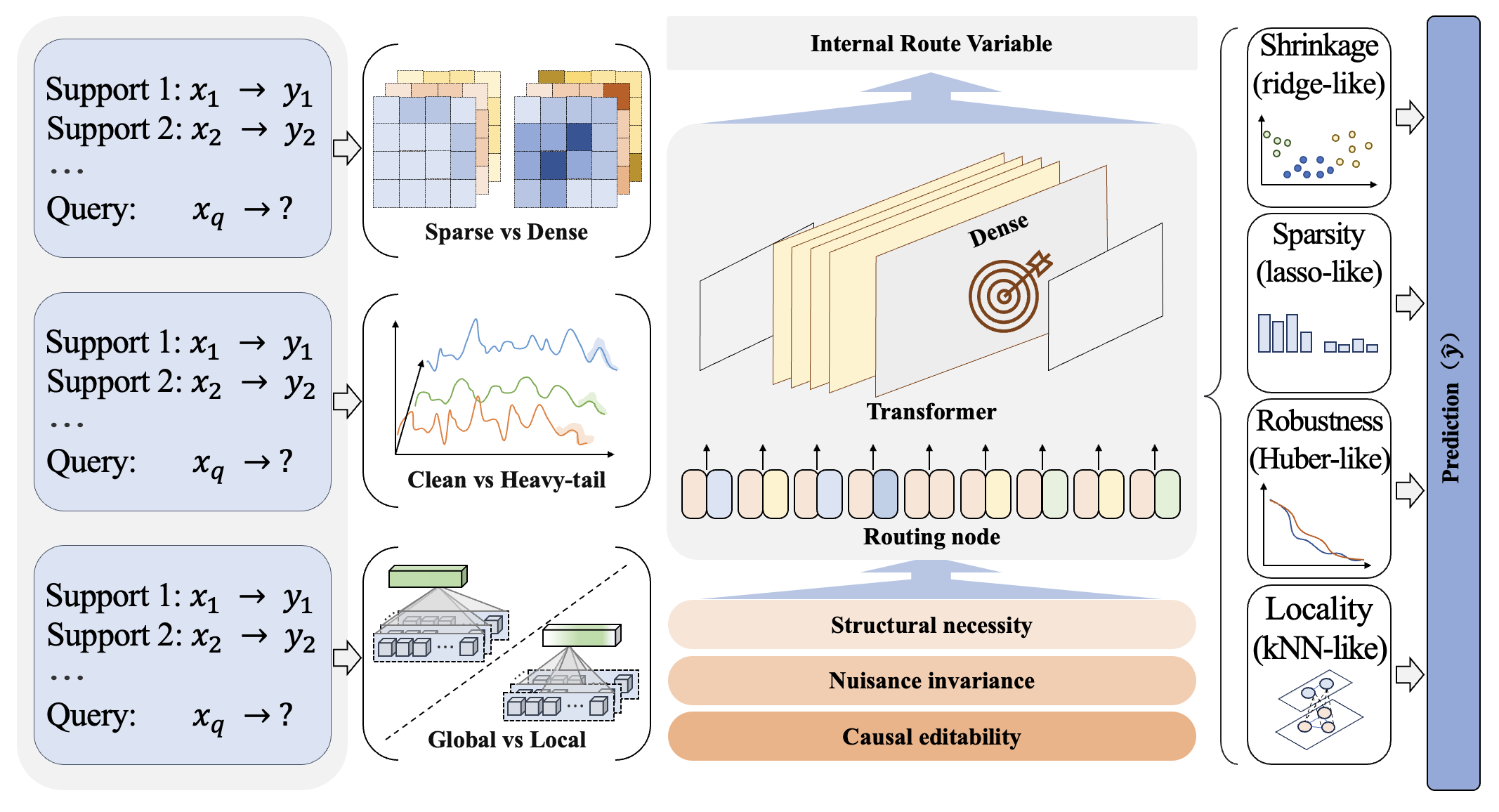}
    \captionsetup{hypcap=false}
    \caption{Grounding latent algorithm routing. The serialized prompt format is held fixed while the latent regime changes. This creates a setting in which inductive-bias routing is structurally necessary while surface shortcuts are discouraged. The paper evaluates route claims against three grounding criteria: structural necessity, nuisance invariance, and causal editability.}
    \label{fig:setup}
    \captionsetup{hypcap=true}
\end{figure}

Figure~\ref{fig:setup} summarizes the logic of the controlled setting. By holding prompt surface form fixed while varying the latent regime, the benchmark creates a setting in which a single fixed solver is structurally disadvantaged, whereas a routed model should satisfy the grounding criteria above. This same setup also motivates the stylized theory below: why fixed family representatives and global soft mixtures are insufficient, and why route variables should be decodable and causally editable in hidden representations.

\subsection{Setup}

Let $z\in\mathcal{Z}$ denote a latent regime and let $R_k(z)$ be the regime-conditional risk of family representative $A_k$:
\[
R_k(z)=\E[\ell(A_k(e),y)\mid z].
\]
The oracle route is
\[
\sigma(z)=\arg\min_{k\in[K]} R_k(z).
\]
We say a predictor uses latent routing if it admits the factorization
\[
e \longmapsto h(e) \longmapsto \route(e) \longmapsto \hat y(e),
\]
where $h(e)$ is an intermediate representation, $\route(e)$ is an example-conditioned route variable, and the final answer $\hat y(e)$ depends on $\route(e)$ through a family-specific downstream computation.

\begin{assumption}[Regime-separated route margins]
For each regime $z$ there exists a unique optimal branch $\sigma(z)$ and a margin $\gamma_z>0$ such that
\[
R_j(z)-R_{\sigma(z)}(z)\ge \gamma_z \qquad \forall j\neq \sigma(z).
\]
Let $\gamma_{\min}=\min_z \gamma_z$.
\end{assumption}

Assumption~1 captures the simplest setting in which routing is structurally necessary: different regimes truly prefer different solver families, and those preferences are separated by a positive risk margin.

\subsection{Why fixed family representatives fail}

The first theorem formalizes the basic task-level reason that routing is needed: mixed-regime problems impose regret on any single fixed family representative.

\begin{theorem}[Unavoidable regret for fixed family representatives]
\label{thm:fixed}
Under Assumption~1, any fixed family representative $A_j$ satisfies
\[
\E_z\!\left[R_j(z)-R_{\sigma(z)}(z)\right]
\ge \Prob(\sigma(z)\neq j)\,\gamma_{\min}.
\]
In particular, if at least two routes occur with prior probability at least $\pi_{\min}>0$, then every fixed family representative incurs expected excess risk at least $\pi_{\min}\gamma_{\min}$.
\end{theorem}

Theorem~\ref{thm:fixed} grounds the need for routing at the task level. A model may still be strong in an absolute sense, but if different regimes induce genuinely different optimal branches, then a single global learner is the wrong inductive object.

\subsection{Why global soft mixtures fail}

A natural alternative to discrete routing is a single soft mixture over solver families. The next theorem shows that this can still be insufficient when regimes induce opposing biases.

\begin{theorem}[Irreducible error of global soft mixtures]
\label{thm:mixture}
Consider two regimes $z\in\{-1,+1\}$ with equal probability and two family representatives $A_+$ and $A_-$. Suppose $A_+$ is unbiased on regime $+1$ and has bias $\Delta$ on regime $-1$, while $A_-$ is unbiased on regime $-1$ and has bias $\Delta$ on regime $+1$. For any global convex mixture
\[
A_\lambda=\lambda A_+ + (1-\lambda)A_-,
\]
the average excess squared error above the discrete oracle router is at least $\Delta^2/4$.
\end{theorem}

The proof is elementary but revealing. Under regime $+1$, the soft mixture retains bias $(1-\lambda)\Delta$; under regime $-1$, it retains bias $\lambda\Delta$. Averaging the squared bias terms gives
\[
\frac{1}{2}\bigl[(1-\lambda)^2+\lambda^2\bigr]\Delta^2 \ge \Delta^2/4.
\]
Thus even the best episode-independent global mixture cannot recover the piecewise structure that a discrete oracle captures exactly. Theorem~\ref{thm:mixture} does not rule out input-conditioned continuous mixtures; we evaluate such alternatives empirically.

\subsection{Why route variables should be readable and editable}

The preceding results explain why routing is useful. The next propositions explain why route variables should also leave identifiable signatures in hidden representations.

\begin{proposition}[Linear decodability of route variables]
\label{prop:decode}
Assume an intermediate statistic vector $\phi(e)\in\R^d$ and route scores
\[
s_k(e)=a_k^\top \phi(e), \qquad \route(e)=\arg\max_k s_k(e),
\]
for vectors $a_k\in\R^d$. If an intermediate representation is
\[
h(e)=M\phi(e)+\xi(e)
\]
with full-column-rank $M$, then in the noiseless case there exists a linear probe that exactly recovers $\route(e)$ from $h(e)$. With additive noise $\xi(e)$, exact recovery still holds whenever the route margin exceeds the induced perturbation:
\[
\min_{j\neq \route(e)} (a_{\route(e)}-a_j)^\top \phi(e)
>
2\|M^+\| \max_k \|a_k\|\,\|\xi(e)\|.
\]
\end{proposition}

A direct consequence is a testable prediction: if intermediate layers encode route-relevant regime statistics before they finish computing the answer, route probes should peak earlier and more sharply than answer probes. Finally, route variables should also be editable when they are represented in a low-dimensional activation subspace.

\begin{proposition}[Sufficient condition for a multiclass route flip]
\label{prop:edit}
Let route scores be $g_k(h)=v_k^\top h$ and let $\route(h)=\arg\max_k g_k(h)$. For a target route $r'$, if an edit vector $\delta$ satisfies
\[
(v_{r'}-v_j)^\top\delta > g_j(h)-g_{r'}(h)
\qquad \forall j\neq r',
\]
then $\route(h+\delta)=r'$. If the downstream readout $D$ is $L$-Lipschitz, then
\[
\|D(h+\delta)-D(h)\|\le L\|\delta\|.
\]
\end{proposition}

Proposition~\ref{prop:edit} gives the conceptual basis for activation patching and steering: route edits can be large enough to cross a route boundary while still small enough to preserve most of the rest of the computation.

\paragraph{Predictions.}
The theory makes three concrete predictions that guide the experiments: (1) mixed-regime tasks should create a large gap between fixed family representatives and routed predictors; (2) route variables should be easier to decode in mid-layer states than final answers; and (3) route-targeted activation edits should change best-fit family behavior while preserving answer quality.

\section{Benchmark and experimental design}

In this section, we describe the benchmark, models, baselines, and evaluation metrics used in our experiments. We first introduce the regime families and their controlled prompt construction, then present the model and baseline setup, and finally define the metrics used to assess both predictive performance and routing behavior.

\subsection{Families and regimes}

\dataset\ contains three pairwise regime contrasts over a
four-family solver bank, with two latent regimes in each contrast.

\paragraph{Sparse vs. dense.}
Support points are drawn from isotropic Gaussian covariates in $\R^{24}$.
The target is linear, but the coefficient vector is either sparse
($\|w\|_0\in[2,4]$) or dense.
Sparse episodes favor the sparsity family, operationalized by lasso-like
behavior, while dense episodes favor the global-shrinkage family,
operationalized by ridge-like behavior
\citep{tibshirani1996lasso,hoerl1970ridge}.

\paragraph{Clean vs. heavy-tailed.}
The underlying predictor remains globally linear, but the noise model changes.
Clean episodes use sub-Gaussian noise; heavy-tailed episodes use a contaminated
distribution with Student-$t$ tails and explicit outliers.
The clean regime favors global-shrinkage estimation, whereas the contaminated
regime favors the robustness family, operationalized by Huber-like behavior
\citep{hoerl1970ridge,huber1964robust}.

\paragraph{Global vs. local.}
Episodes are generated either from a globally linear predictor or from a local
piecewise-smooth rule defined by cluster-conditioned slopes.
The local regime favors the locality family, operationalized by $k$NN-like
matching, while the global regime again favors global-shrinkage aggregation
\citep{cover1967nearest,hoerl1970ridge}.

For all regime contrasts, query serialization, separators, and numeric
formatting are randomized independently of the regime.

\subsection{Models and baselines}

We train decoder-only transformers with 44M, 89M, 167M, 306M, and 612M parameters on mixed-regime next-token prediction only.
No route labels, branch identifiers, or chain-of-thought traces are provided.
All models use the same tokenizer, 1024-token context budget, cosine decay schedule, and contrast-balanced minibatching.

We compare against the following alternatives. \textbf{Fixed representatives} instantiate shrinkage, sparsity, robustness, and locality. A \textbf{global soft mixture} learns one episode-independent convex combination of representative outputs. An \textbf{input-conditioned soft mixture} predicts episode-dependent softmax weights from a learned summary of the support/query episode and is trained only with answer loss. An \textbf{unsupervised Gumbel router} uses a straight-through Gumbel--Softmax gate over the same solver bank, also without route labels. We use \textbf{privileged-stat router} for the baseline previously called the external router. It receives a fixed vector of hand-computed support/query summaries chosen to expose the controlled statistical axes varied by \dataset---sparsity/concentration, tail/outlier structure, and local-versus-global regularity---but it receives neither the discrete latent route label nor transformer activations. It is therefore a diagnostic privileged baseline rather than a deployable competitor. This is distinct from the \textbf{oracle router}, which dispatches directly from the true latent regime. Appendix~\ref{sec:router-ablation} separates solver-output, episode-stat, combined, privileged-stat, and true-oracle variants.

\subsection{Metrics}

% Our main scalar accuracy metric is normalized RMSE (NRMSE).
% To quantify routing, we compute route F1 between the behavioral route $\hat \route_{\mathrm{beh}}$ and the oracle route $\sigma(z)$.
% We additionally report:
% \begin{itemize}[leftmargin=1.0em]
% \item \textbf{Oracle gap closed}: improvement over the best fixed family representative as a fraction of the gap to the oracle router.
% \item \textbf{Nuisance consistency}: percentage of formatting-only perturbations that preserve $\hat \route_{\mathrm{beh}}$.
% \item \textbf{True-regime sensitivity}: percentage of matched-pair regime swaps that change $\hat \route_{\mathrm{beh}}$ in the expected direction.
% \end{itemize}

Our main scalar accuracy metric is normalized RMSE (NRMSE).
\textbf{Route F1} is macro-F1 between the behavioral proxy
$\hat\route_{\mathrm{beh}}$ and the oracle route $\sigma(z)$.
For a method with NRMSE $E_{\mathrm{method}}$, we define
\textbf{oracle gap closed} as
\[
\mathrm{GapClosed}
=
100\times
\frac{E_{\mathrm{fixed}}-E_{\mathrm{method}}}
     {E_{\mathrm{fixed}}-E_{\mathrm{oracle}}},
\]
where $E_{\mathrm{fixed}}$ is the NRMSE of the best fixed representative
and $E_{\mathrm{oracle}}$ is the NRMSE of the oracle router.
\textbf{Nuisance consistency} is the percentage of semantics-preserving
perturbation pairs that preserve $\hat\route_{\mathrm{beh}}$, and
\textbf{true-regime sensitivity} is the percentage of matched regime swaps
that change it in the expected direction.

For interventions, an episode is \emph{eligible} when its pre-edit behavioral route differs from the target family. \textbf{Target flip} is the percentage of eligible episodes whose post-edit route equals that target; \textbf{non-target flip} is the percentage whose post-edit route changes to any other non-target family. \textbf{Answer retention} is the post-intervention answer-quality score normalized by the corresponding original, unmodified model score and
reported as a percentage, with 100 indicating no loss relative to the
original model output. It is evaluated on the same eligible intervention episodes used for the flip metrics. We report the absolute \textbf{NRMSE degradation} $\Delta\mathrm{NRMSE}$ separately; retention and $\Delta\mathrm{NRMSE}$ are complementary evaluation quantities and are not algebraically derived from one another.

\section{Main empirical results}

\begin{table}[!htbp]
\centering
\small
\setlength{\tabcolsep}{3.8pt}
\renewcommand{\arraystretch}{1.04}
\resizebox{\columnwidth}{!}{%
\begin{tabular}{lcccccc}
\toprule
Method & ID $\downarrow$ & OOD $\downarrow$ & F1 $\uparrow$
& Gap $\uparrow$ & Cons. $\uparrow$ & Sens. $\uparrow$\\
\midrule
Best fixed family & 0.239 & 0.293 & 48.2 & 0.0 & 74.1 & 43.5\\
Global soft mixture & 0.219 & 0.255 & 59.7 & 22.5 & 83.5 & 61.0\\
Input-conditioned mixture & 0.184 & 0.217 & 72.8 & 61.8 & 89.6 & 74.9\\
Unsupervised Gumbel router & 0.176 & 0.207 & 77.9 & 70.8 & 91.8 & 79.6\\
167M Transformer & 0.182 & 0.213 & 77.6 & 64.0 & 93.7 & 81.4\\
306M Transformer & 0.167 & 0.194 & 84.1 & 80.9 & 94.8 & 85.9\\
612M Transformer & 0.162 & 0.189 & 85.8 & 86.5 & 95.2 & 87.1\\
Privileged-stat router & 0.165 & 0.192 & 87.5 & 83.1 & 91.2 & 84.4\\
Oracle router & 0.150 & 0.174 & 100.0 & 100.0 & 100.0 & 100.0\\
\bottomrule
\end{tabular}}
\caption{Aggregate comparison with adaptive alternatives. ID and OOD
NRMSE are macro-averaged over the three pairwise regime contrasts.
The input-conditioned mixture and Gumbel router use no route labels.
The privileged-stat router uses hand-computed episode statistics;
only the oracle router receives the true route.}
\label{tab:adaptive}
\end{table}

\begin{table}[t]
\centering
\small
\setlength{\tabcolsep}{3.8pt}
\renewcommand{\arraystretch}{1.05}
\resizebox{\textwidth}{!}{%
\begin{tabular}{lcccccccccccccc}
\toprule
& \multicolumn{3}{c}{Sparse / dense} & \multicolumn{3}{c}{Heavy-tail} & \multicolumn{3}{c}{Local / global} & \multicolumn{4}{c}{Aggregate} \\
\cmidrule(lr){2-4} \cmidrule(lr){5-7} \cmidrule(lr){8-10} \cmidrule(lr){11-14}
Method & ID $\downarrow$ & OOD $\downarrow$ & Route F1 $\uparrow$ & ID $\downarrow$ & OOD $\downarrow$ & Route F1 $\uparrow$ & ID $\downarrow$ & OOD $\downarrow$ & Route F1 $\uparrow$ & Avg. ID $\downarrow$ & Avg. F1 $\uparrow$ & Gap closed $\uparrow$ & Consistency $\uparrow$ \\
\midrule
Shrinkage family & 0.262 & 0.301 & 48.2 & 0.236 & 0.334 & 45.8 & 0.219 & 0.245 & 50.7 & 0.239 & 48.2 & -- & 74.1 \\
Sparsity family & 0.241 & 0.288 & 44.6 & 0.311 & 0.366 & 10.9 & 0.298 & 0.351 & 12.8 & 0.283 & 22.8 & -- & 68.0 \\
Robustness family & 0.279 & 0.319 & 15.7 & 0.207 & 0.254 & 49.6 & 0.271 & 0.319 & 18.4 & 0.252 & 27.9 & -- & 71.8 \\
Locality family & 0.301 & 0.347 & 9.8 & 0.338 & 0.391 & 7.1 & 0.198 & 0.229 & 47.9 & 0.279 & 21.6 & -- & 66.2 \\
Soft mixture & 0.213 & 0.249 & 60.9 & 0.228 & 0.264 & 58.1 & 0.217 & 0.253 & 60.1 & 0.219 & 59.7 & 22.5 & 83.5 \\
\rowcolor{softgray} Privileged-stat router & 0.158 & 0.187 & 88.1 & 0.171 & 0.199 & 86.9 & 0.165 & 0.190 & 87.4 & 0.165 & 87.5 & 83.1 & 91.2 \\
44M & 0.230 & 0.268 & 57.8 & 0.244 & 0.280 & 55.0 & 0.229 & 0.263 & 56.3 & 0.234 & 56.4 & 5.6 & 84.4 \\
89M & 0.208 & 0.243 & 63.8 & 0.223 & 0.258 & 61.5 & 0.212 & 0.248 & 64.2 & 0.214 & 63.2 & 28.1 & 88.9 \\
167M & 0.176 & 0.209 & 78.4 & 0.189 & 0.220 & 75.2 & 0.181 & 0.211 & 79.1 & 0.182 & 77.6 & 64.0 & 93.7 \\
306M & 0.162 & 0.189 & 84.7 & 0.173 & 0.200 & 82.4 & 0.166 & 0.192 & 85.3 & 0.167 & 84.1 & 80.9 & 94.8 \\
\underline{612M} & 0.157 & 0.184 & 86.3 & 0.169 & 0.196 & 84.1 & 0.160 & 0.186 & 86.9 & 0.162 & 85.8 & 86.5 & 95.2 \\
\rowcolor{softgray} \textbf{Oracle router} & 0.142 & 0.168 & 100.0 & 0.159 & 0.182 & 100.0 & 0.148 & 0.171 & 100.0 & 0.150 & 100.0 & 100.0 & 100.0 \\
\bottomrule
\end{tabular}}
\caption{Main benchmark results on \dataset. ID and OOD denote in-distribution and held-out perturbation splits. The four family rows correspond to fixed shrinkage, sparsity, robustness, and locality representatives, instantiated by ridge-like, lasso-like, Huber-like, and $k$NN-like predictors.}
\label{tab:main}
\end{table}

Tables~\ref{tab:adaptive} and~\ref{tab:main} give the central empirical comparison. Fixed representatives are strong only in aligned regimes. The global mixture improves average fit but has route F1 59.7. The input-conditioned mixture and Gumbel router are substantially stronger, validating the continuous-adaptation alternative, yet the 306M/612M transformers retain higher route F1, better OOD NRMSE, and higher nuisance consistency. The 306M model closes 80.9\% of the oracle gap with route F1 84.1, while the 612M model closes 86.5\% and reaches 95.2 consistency.

The privileged-stat router remains slightly stronger in route F1 because it receives hand-computed regime statistics. It is not an oracle: it receives neither the true route label nor model activations. The true oracle router is reported separately. This separation makes the empirical interpretation conservative: dense route-like behavior approaches a privileged diagnostic baseline while retaining a measurable gap to direct latent-route access.

\begin{figure}[!htbp]
    \centering
    % Regenerate this source figure with the legend label
% ``Privileged-stat router'' (or ``Priv.-stat router'' if space is limited).
\includegraphics[width=0.99\textwidth]{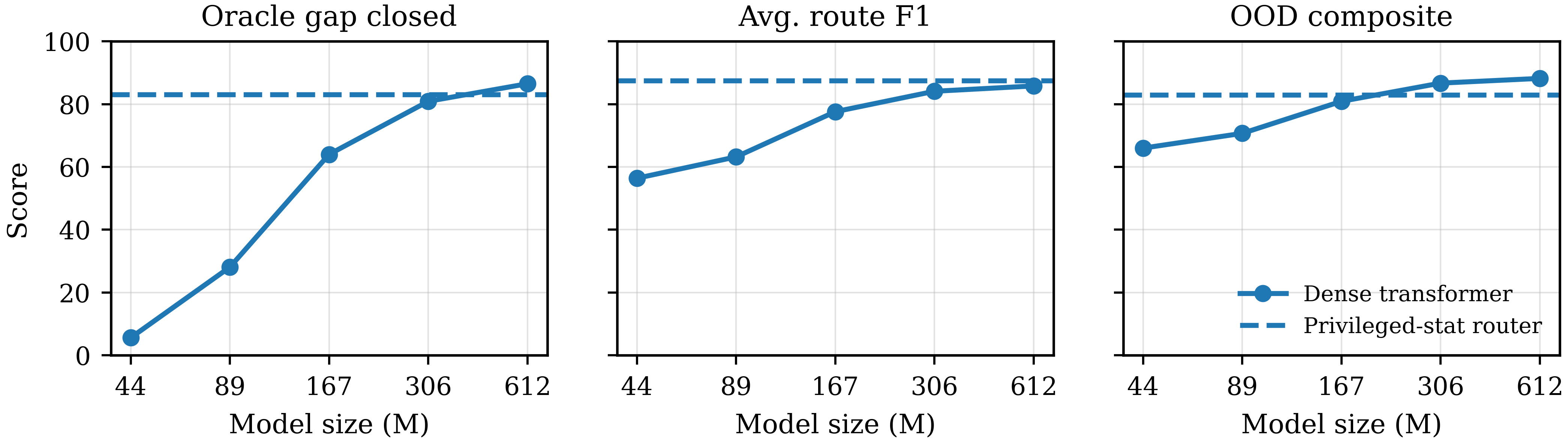}
    \caption{Scaling trends across three grounding metrics.
    Left: oracle-gap closure.
    Middle: average route F1.
    Right: OOD composite.
    Breaking the view into matched panels makes the coupled scaling trend easier to compare, while the dashed reference line in each panel shows the privileged-stat-router baseline.}
    \label{fig:scaling}
\end{figure}

Figure~\ref{fig:scaling} shows that scale strengthens all three grounding dimensions together.
If route-like behavior were only a side effect of prompt memorization, one would expect route F1 and nuisance robustness to decouple from the accuracy curve.
Instead they move in tandem.

\begin{table}[!htbp]
\centering
\small
\setlength{\tabcolsep}{4.2pt}
\renewcommand{\arraystretch}{1.05}
\resizebox{\textwidth}{!}{%
\begin{tabular}{lccccccc}
\toprule
Method & Format shift $\Delta$NRMSE $\downarrow$ & Order shift $\Delta$NRMSE $\downarrow$ & Lexical rewrite $\Delta$NRMSE $\downarrow$ & Prior-shift gap closed $\uparrow$ & Nuisance consistency $\uparrow$ & True-regime sensitivity $\uparrow$ & OOD composite $\uparrow$ \\
\midrule
Soft mixture & 0.041 & 0.036 & 0.039 & 57.2 & 83.5 & 61.0 & 62.4 \\
\rowcolor{softgray} Privileged-stat router & 0.028 & 0.024 & 0.030 & 82.7 & 91.2 & 84.4 & 83.0 \\
89M & 0.024 & 0.021 & 0.026 & 64.8 & 88.9 & 71.8 & 70.7 \\
167M & 0.019 & 0.017 & 0.021 & 78.3 & 93.7 & 81.4 & 80.9 \\
306M & 0.016 & 0.015 & 0.018 & 86.1 & 94.8 & 85.9 & 86.7 \\
\underline{612M} & 0.015 & 0.014 & 0.017 & 88.0 & 95.2 & 87.1 & 88.2 \\
\bottomrule
\end{tabular}}
\caption{Grounding under nuisance and regime perturbations.
Small degradation under formatting-only perturbations is necessary but not sufficient: a grounded routed model must also react strongly to true latent regime changes.
The 167M+ models are better than the soft mixture on both sides of this tradeoff.}
\label{tab:ood}
\end{table}

Table~\ref{tab:ood} makes the grounding criterion explicit.
A model could be highly format-invariant yet insensitive to real regime changes, or very sensitive yet also brittle to nuisance perturbations.
The routed-transformer regime is characterized by doing well on both.
Relative to the soft mixture, the 167M model nearly halves the damage from formatting perturbations and improves true-regime sensitivity from 61.0 to 81.4.
The larger models continue this trend.
These results matter because they distinguish regime response from simple style matching.

\begin{table}[!htbp]
\centering
\small
\setlength{\tabcolsep}{2.8pt}
\renewcommand{\arraystretch}{0.94}

\begin{minipage}[t]{0.47\textwidth}
\centering
\resizebox{\linewidth}{!}{%
\begin{tabular}{lccccc}
\toprule
Rendering & ID $\downarrow$ & OOD $\downarrow$ & F1 $\uparrow$ &
Cons. $\uparrow$ & Sens. $\uparrow$\\
\midrule
Numeric & 0.167 & 0.194 & 84.1 & 94.8 & 85.9\\
Natural language & 0.174 & 0.203 & 81.6 & 92.9 & 83.2\\
NL + shuffle & 0.178 & 0.207 & 80.4 & 91.8 & 82.1\\
NL + paraphrase & 0.181 & 0.211 & 79.7 & 91.1 & 81.5\\
\bottomrule
\end{tabular}}
\captionof{table}{Rendering robustness for the 306M model.}
\label{tab:nl}
\end{minipage}
\hfill
\begin{minipage}[t]{0.50\textwidth}
\centering
\resizebox{\linewidth}{!}{%
\begin{tabular}{lccccc}
\toprule
Method & ID $\downarrow$ & OOD $\downarrow$ & F1 $\uparrow$ &
Gap $\uparrow$ & Cons. $\uparrow$\\
\midrule
Best fixed & 0.268 & 0.319 & 35.4 & 0.0 & 70.8\\
Global mixture & 0.239 & 0.281 & 46.7 & 25.2 & 80.1\\
Cond. mixture & 0.204 & 0.243 & 63.5 & 55.7 & 86.7\\
Gumbel router & 0.195 & 0.231 & 68.9 & 63.5 & 88.4\\
306M & 0.181 & 0.214 & 76.8 & 75.7 & 92.7\\
612M & 0.176 & 0.207 & 79.4 & 80.0 & 93.6\\
Priv.-stat & 0.174 & 0.205 & 82.6 & 81.7 & 90.4\\
Oracle & 0.153 & 0.181 & 100.0 & 100.0 & 100.0\\
\bottomrule
\end{tabular}}
\captionof{table}{Unified four-way \dataset.}
\label{tab:fourway}
\end{minipage}
\end{table}

Table~\ref{tab:nl} shows that the route signal weakens moderately but remains substantial under natural-language rendering, support shuffling, and lexical paraphrase. Table~\ref{tab:fourway} shows the same ordering when all four families compete simultaneously, ruling out an explanation tied only to binary regime contrasts.

\section{Internal evidence for route variables}

Behavioral matching is necessary but not sufficient for a mechanistic interpretation. We therefore test whether route-relevant information is decodable, whether probes survive nuisance and label controls, and whether targeted edits change solver-family-consistent behavior more selectively than generic activation corruption.

For activation patching, we construct matched source/target episodes with different oracle routes while controlling prompt format. At the route-probe peak layer, we cache residual-stream activations, project onto the identified route subspace, replace the corresponding target component with the matched source component, and decode the patched target. We report target flip, non-target flip, answer retention, and NRMSE degradation. Random-source, same-route, and wrong-route patches serve as controls. The analysis follows standard matched-pair probing and activation-patching methodology \citep{vig2020causal,meng2022rome,goldowsky2023path,zhang2024patching,heimersheim2024patching}.

\begin{figure}[!htbp]
    \centering
    \includegraphics[width=0.96\textwidth]{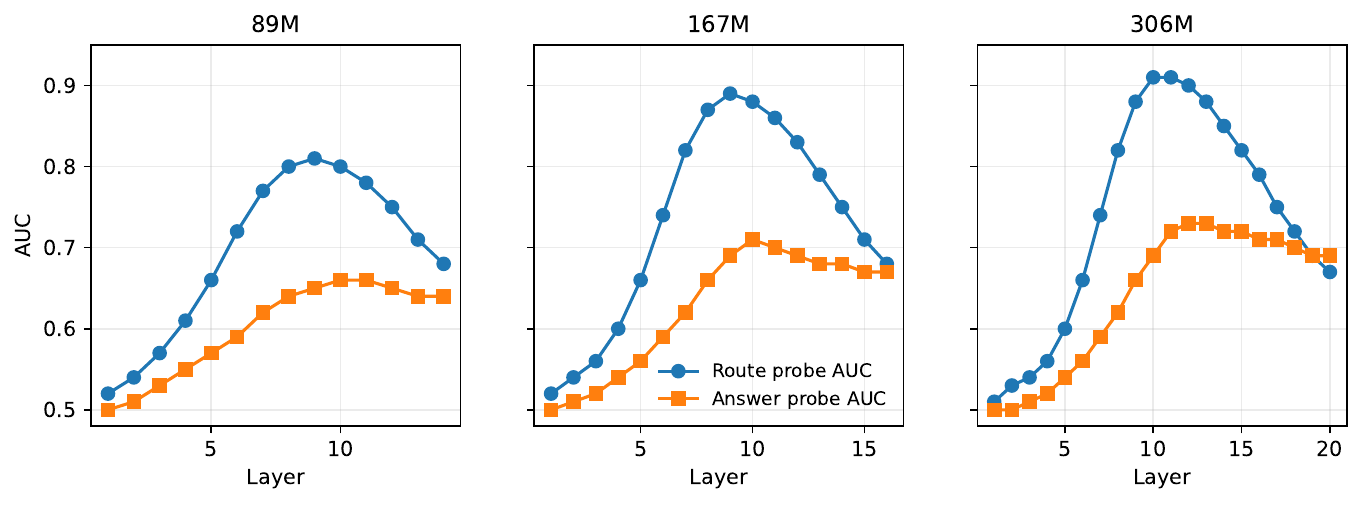}
    \caption{Layerwise route and answer probes. Route probes peak in middle layers and earlier than answer probes, consistent with Proposition~\ref{prop:decode}.}
    \label{fig:layer}
\end{figure}

Figure~\ref{fig:layer} shows that route probes peak earlier than answer probes, reaching AUC 0.89 and 0.91 for the 167M and 306M models, consistent with route-relevant statistics being encoded before final answer computation.

\begin{table}[!htbp]
\centering
\small
\setlength{\tabcolsep}{3.3pt}
\renewcommand{\arraystretch}{1.05}
\resizebox{\columnwidth}{!}{%
\begin{tabular}{lccccccc}
\toprule
Model & Peak route AUC $\uparrow$ & Peak answer AUC $\uparrow$ & Earliest 0.75 route layer $\downarrow$ & Patch flip $\uparrow$ & Steering shift $\uparrow$ & Answer retention $\uparrow$ & Consistency $\uparrow$ \\
\midrule
89M & 0.81 & 0.65 & 9.6 & 48.3 & 19.4 & 94.6 & 88.9 \\
167M & 0.89 & 0.71 & 7.8 & 69.1 & 31.6 & 96.0 & 93.7 \\
306M & 0.91 & 0.73 & 8.1 & 73.8 & 35.9 & 96.7 & 94.8 \\
\underline{612M} & 0.92 & 0.74 & 8.9 & 75.2 & 37.1 & 97.1 & 95.2 \\
\bottomrule
\end{tabular}}
\caption{Representation and intervention evidence.
Patch flip is the target-family behavioral flip rate over eligible matched-pair episodes.
Steering shift is the increase in target-family share under route-direction steering.
Answer retention is post-intervention answer quality normalized by the
corresponding original, unmodified model score.
Consistency is behavioral-route consistency under semantics-preserving nuisance perturbations.}
\label{tab:internal}
\end{table}

Table~\ref{tab:internal} shows that matched target-route patching yields a 73.8\% target-family behavioral flip rate with 96.7\% answer retention for the 306M model. Random-source, same-route, and wrong-route patches are substantially less target-selective. This pattern supports functional involvement of the identified route-relevant directions without implying a discrete symbolic router.

\begin{figure}[!htbp]
    \centering
    \includegraphics[width=0.99\textwidth]{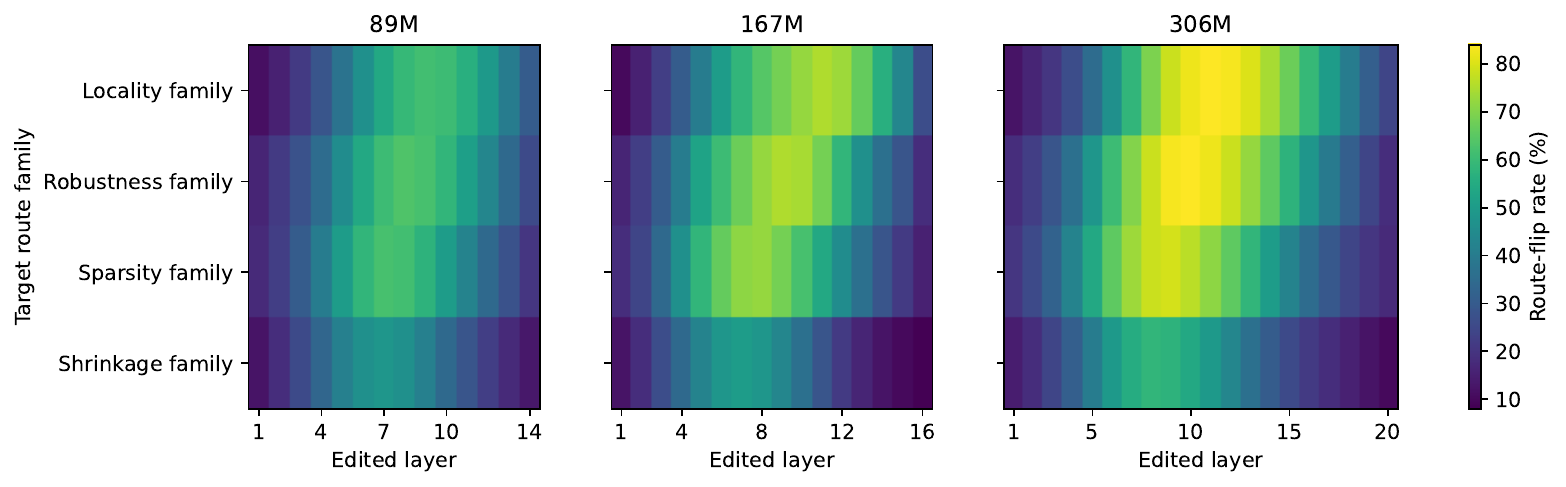}
    \caption{Matched-pair activation patching. Route-flip rates concentrate in middle-layer residual states and become more localized at larger scales, overlapping with the route-probe peaks.}
    \label{fig:patch}
\end{figure}

\section{Conclusion}
\label{sec:conclusion}
We studied whether dense decoder-only transformers trained from scratch on a controlled mixed-regime benchmark can develop route-like internal computation over latent inductive-bias families. \dataset\ evaluates this interpretation through structural necessity, nuisance invariance, and causal editability while treating closest-solver identity explicitly as a behavioral proxy.

Across 44M--612M models, route F1, oracle-gap closure, and nuisance robustness improve with scale. The signal persists under natural-language renderings and a unified four-way setting, and larger dense models outperform input-conditioned soft-mixture and unsupervised Gumbel-routing alternatives. Probe permutation controls, margin-filtered analyses, and matched patching controls indicate that route-relevant information is not merely a formatting artifact or arbitrary post-hoc label.

The scope is deliberately limited. These findings provide controlled evidence that dense transformers \emph{can} organize computation around route-like internal variables on \dataset; they do not establish the same mechanism in pretrained LLMs, natural-language QA, math, code generation, or unrestricted agentic reasoning. Extending the diagnostic to those settings, and more fully characterizing how latent activation-level routing relates to explicit expert routing across architectures, scales, and naturalistic tasks, remain open directions.

\bibliographystyle{colm2026_conference}
\bibliography{COLM}

\appendix

\section{Proof sketches and additional theoretical detail}

\subsection{Proof of Theorem~\ref{thm:fixed}}

Fix a family representative $A_j$.
For any regime $z$ with $\sigma(z)=j$, the regret term is zero.
For any regime with $\sigma(z)\neq j$, Assumption~1 gives
\[
R_j(z)-R_{\sigma(z)}(z)\ge \gamma_{\min}.
\]
Taking expectation over $z$ immediately yields
\[
\E_z[R_j(z)-R_{\sigma(z)}(z)]
\ge \Prob(\sigma(z)\neq j)\gamma_{\min}.
\]
The lower bound is informative whenever more than one route appears with nontrivial mass.
In that case, no global fixed family representative can match the routed oracle even if it is optimal on one regime.

\subsection{Proof of Theorem~\ref{thm:mixture}}

Under regime $+1$, the convex mixture $A_\lambda$ retains bias $(1-\lambda)\Delta$ because the $A_-$ component is wrong by $\Delta$.
Under regime $-1$, the retained bias is $\lambda \Delta$.
If the oracle router is unbiased in both regimes, then the excess squared error of the soft mixture is
\[
\frac{1}{2}(1-\lambda)^2\Delta^2 + \frac{1}{2}\lambda^2\Delta^2
= \frac{\Delta^2}{2}\bigl((1-\lambda)^2+\lambda^2\bigr).
\]
The quadratic term is minimized at $\lambda=\nicefrac{1}{2}$, giving minimum excess error $\Delta^2/4$.
Thus any global soft mixture leaves irreducible bias whenever the two regimes require opposing inductive-bias families.

\subsection{Proof sketch of Proposition~\ref{prop:decode}}

Because $M$ has full column rank, its pseudoinverse satisfies $M^+M=I$.
Define $b_k=(M^+)^\top a_k$.
Then, in the noiseless case,
\[
b_k^\top h(e)=a_k^\top M^+M\phi(e)=a_k^\top \phi(e)=s_k(e),
\]
so a linear probe over $h(e)$ recovers the route exactly.
With additive noise, the perturbation in each route score is bounded by
\[
|b_k^\top \xi(e)| \le \|b_k\|\,\|\xi(e)\|
\le \|M^+\|\|a_k\|\,\|\xi(e)\|.
\]
A standard margin argument yields the stated condition for exact recovery.
The empirical implication is that route probes should perform well whenever the residual stream preserves a sufficiently linear image of regime statistics.

\subsection{Proof sketch of Proposition~\ref{prop:edit}}

For every competitor $j\neq r'$, the stated condition gives
\[
g_{r'}(h+\delta)-g_j(h+\delta)
= g_{r'}(h)-g_j(h)+(v_{r'}-v_j)^\top\delta > 0.
\]
Hence $g_{r'}(h+\delta)$ exceeds every competing score and $\route(h+\delta)=r'$. If the downstream map $D$ is $L$-Lipschitz, the induced answer shift is at most $L\|\delta\|$. The condition is sufficient rather than necessary; its role is to show that a multiclass route boundary can be crossed without requiring an arbitrarily large perturbation to the full hidden state.

\subsection{Further interpretation}

Theorems~\ref{thm:fixed}--\ref{thm:mixture} and Propositions~\ref{prop:decode}--\ref{prop:edit} do not attempt to model every detail of transformer computation.
Their role is to clarify what kinds of signatures one should expect if latent routing is the right explanatory object.
In particular, they explain why the main empirical phenomena should come as a package:
large fixed-family-representative regret, improvement from discrete routing over soft interpolation, early decodability of route variables, and successful route-targeted interventions.

\section{Additional benchmark details}

Each pairwise regime contrast contributes equally to the training
mixture. Within each contrast, the two latent regimes are sampled
with equal probability.
Support size is 32 in the main benchmark, query dimension is 24, and numeric values are serialized using randomized decimal precision and separator templates to reduce shortcutting on lexical surface cues.
The OOD suite includes independent perturbations of formatting, support order, lexical wrappers, and family priors, plus matched-pair counterfactual regime swaps.

All transformer models are trained with AdamW, cosine learning-rate decay, and a warmup phase of 2k updates.
The 44M, 89M, 167M, 306M, and 612M models scale width, depth, and feed-forward multiplier jointly while keeping the tokenizer and context window fixed.
Validation-based early stopping is disabled; all models train for the same token budget to preserve the scaling comparison.
\section{Additional experimental results}

\subsection{Router ablations}
\label{sec:router-ablation}

The baseline previously denoted ``external router'' is the privileged-stat
router below. It consumes a fixed vector of hand-computed summaries of the
observed support/query episode selected to expose the benchmark's controlled
axes: sparsity/concentration, tail/outlier structure, and local-versus-global
regularity. It receives no transformer activation, discrete latent route label,
or branch identifier. Because these summaries are explicitly engineered around
the \dataset\ construction, this baseline is used only as a privileged
diagnostic reference. The true oracle, which receives the latent route itself,
is listed separately.

\begin{table}[!htbp]
\centering
\small
\setlength{\tabcolsep}{4.0pt}
\resizebox{\textwidth}{!}{%
\begin{tabular}{lllcccc}
\toprule
Router & Inputs & Route labels?
& ID $\downarrow$
& OOD $\downarrow$
& F1 $\uparrow$
& Gap $\uparrow$ \\
\midrule
Solver-output gate
& Solver outputs
& No
& 0.197
& 0.231
& 66.4
& 47.2 \\

Episode-stat gate
& Support/query statistics
& No
& 0.188
& 0.222
& 70.8
& 57.3 \\

Solver + stat gate
& Outputs + statistics
& No
& 0.181
& 0.214
& 74.6
& 65.2 \\

Privileged-stat router
& Privileged statistics
& No
& 0.165
& 0.192
& 87.5
& 83.1 \\
\bottomrule
\end{tabular}%
}
\caption{
Router ablations. Privileged statistics create a diagnostic upper baseline
without revealing the true route label.
}
\label{tab:router-ablation}
\end{table}

\subsection{Route-proxy, probe, and patching controls}
\label{sec:route-controls}

\begin{table}[!htbp]
\centering
\small
\setlength{\tabcolsep}{3.7pt}
\resizebox{\textwidth}{!}{%
\begin{tabular}{lcccccc}
\toprule
Subset
& Coverage
& F1 $\uparrow$
& Oracle agreement $\uparrow$
& Probe AUC $\uparrow$
& Patch flip $\uparrow$
& Retention $\uparrow$ \\
\midrule
All episodes
& 100\%
& 84.1
& 81.7
& 0.91
& 73.8
& 96.7 \\
Top 75\% margin
& 75\%
& 87.6
& 85.4
& 0.93
& 76.9
& 96.5 \\
Top 50\% margin
& 50\%
& 91.2
& 89.8
& 0.95
& 81.7
& 96.1 \\
Top 25\% margin
& 25\%
& 94.8
& 93.6
& 0.96
& 85.9
& 95.4 \\
Bottom 25\% margin
& 25\%
& 66.3
& 62.9
& 0.78
& 51.2
& 97.2 \\
\bottomrule
\end{tabular}%
}
\caption{
Margin-filtered sanity check for the behavioral route proxy. Higher solver
margins yield stronger oracle agreement, probe AUC, and targeted patching
effects.
}
\label{tab:margin}
\end{table}

\begin{table}[!htbp]
\centering
\small
\setlength{\tabcolsep}{4.0pt}
\begin{tabular}{lccp{0.42\textwidth}}
\toprule
Probe target
& Peak AUC $\uparrow$
& Earliest 0.75 layer $\downarrow$
& Interpretation \\
\midrule
True route label
& 0.91
& 8.1
& Route-relevant information is recoverable \\
Permuted route label
& 0.51
& --
& No random-label recovery \\
Formatting template ID
& 0.56
& --
& Probe is not primarily reading format \\
Episode ID bucket
& 0.52
& --
& Probe is not memorizing episode identity \\
Solver-margin bucket
& 0.73
& 10.4
& Margin is encoded, but less strongly and later \\
\bottomrule
\end{tabular}
\caption{
Probe permutation and nuisance controls for the 306M model.
}
\label{tab:probe-controls}
\end{table}

\paragraph{Intervention metrics.}
For every eligible episode, target flip and non-target flip are computed from
the post-intervention behavioral route. Answer retention is the intervention
evaluator's normalized answer-quality score divided by the corresponding
original, unmodified model score, expressed as a percentage. The absolute
increase in NRMSE is reported separately as $\Delta\mathrm{NRMSE}$. Because
retention and NRMSE summarize different aspects of answer preservation, the
reported retention values are not computed from $\Delta\mathrm{NRMSE}$ by a
fixed algebraic conversion. Table~\ref{tab:patch-controls} compares matched
target-route patching against random-source, same-route, and wrong-route
activation-replacement controls.

\begin{table}[!htbp]
\centering
\footnotesize
\setlength{\tabcolsep}{2.6pt}
\renewcommand{\arraystretch}{0.96}
\resizebox{\columnwidth}{!}{%
\begin{tabular}{lrrrr}
\toprule
Patch condition
& \shortstack{Target flip\\$\uparrow$}
& \shortstack{Non-target flip\\$\downarrow$}
& \shortstack{Retention\\$\uparrow$}
& \shortstack{NRMSE deg.\\$\downarrow$} \\
\midrule
Random-source patch
& 18.7
& 21.5
& 92.8
& 0.031 \\
Same-route patch
& 13.2
& 10.6
& 97.4
& 0.008 \\
Wrong-route patch
& 24.5
& 39.8
& 91.9
& 0.037 \\
Matched target-route patch
& 73.8
& 9.7
& 96.7
& 0.012 \\
\bottomrule
\end{tabular}%
}
\caption{
Activation-patching controls for the 306M model. Target and non-target flip
rates are computed over eligible episodes whose pre-intervention behavioral
route differs from the requested target. Answer retention and absolute NRMSE
degradation are reported separately. Matched target-route patching produces
a substantially more target-selective behavioral shift than the random-source,
same-route, and wrong-route controls.
}
\label{tab:patch-controls}
\end{table}

Matched target-route patching yields a substantially larger target-family
behavioral flip rate than all three controls, while producing less non-target
switching than the wrong-route control. This pattern is more consistent with
a target-selective intervention than with generic activation corruption; it
does not establish a discrete symbolic router.

\subsection{Explicit routing diagnostic}
\label{sec:moe}

We compare dense models with Switch-MoE variants to test how latent
activation-level routing relates to explicit expert routing. The labels
``167M-equivalent'' and ``306M-equivalent'' refer to the matched backbone-scale
configurations used in the diagnostic. They indicate correspondence to the
dense model's nominal scale and training setup, not exact equality in total
parameter count, active parameters per token, or FLOPs. The top-1 and top-2
variants use the same nominal backbone scale but activate one and two experts
per routed token, respectively, so the top-2 configuration has greater active
computation. Expert purity measures the fraction of query-position expert
assignments consistent with each expert's majority oracle solver family.
Expert MI is the mutual-information score between query-position expert
identity and oracle solver-family identity, computed with the same
normalization across all MoE variants.

\begin{table}[!htbp]
\centering
\small
\setlength{\tabcolsep}{4.0pt}
\resizebox{\textwidth}{!}{%
\begin{tabular}{lccccc}
\toprule
Model
& ID $\downarrow$
& F1 $\uparrow$
& Expert purity $\uparrow$
& Expert MI $\uparrow$
& Cons. $\uparrow$ \\
\midrule
Dense 167M
& 0.182
& 77.6
& --
& --
& 93.7 \\
Dense 306M
& 0.167
& 84.1
& --
& --
& 94.8 \\
Switch-MoE 167M-equivalent, top-1
& 0.174
& 80.3
& 71.6
& 0.42
& 91.5 \\
Switch-MoE 306M-equivalent, top-1
& 0.161
& 85.2
& 76.9
& 0.49
& 92.4 \\
Switch-MoE 306M-equivalent, top-2
& 0.158
& 86.0
& 73.4
& 0.46
& 93.1 \\
\bottomrule
\end{tabular}%
}
\caption{
Switch-MoE diagnostic. The ``equivalent'' labels denote matched nominal
backbone-scale configurations and should not be interpreted as exact
total-parameter or FLOP matching. Expert purity and expert MI are computed
between query-position expert selection and the oracle solver-family identity.
Explicit routing improves ID NRMSE and route F1, but expert identity is only
partially aligned with solver-family identity and does not automatically
improve nuisance consistency.
}
\label{tab:moe}
\end{table}

\paragraph{Per-regime behavior.}
Table~\ref{tab:perregime} shows why fixed family representatives are
structurally limited on \dataset. Each representative is competitive only on
the regimes aligned with its inductive bias: the sparsity family is best among
fixed solvers on sparse structure, the robustness family is best on heavy-tail
noise, and the locality family is best on local signal. Outside those aligned
regimes, the same representatives degrade sharply. By contrast, the 167M
transformer remains much closer to the oracle across all six regimes, which is
the pattern expected from latent routing rather than commitment to a single
global solver.

\begin{table}[!htbp]
\centering
\small
\setlength{\tabcolsep}{4.2pt}
\renewcommand{\arraystretch}{1.05}
\resizebox{\textwidth}{!}{%
\begin{tabular}{llccccccc}
\toprule
Latent regime
& Preferred family
& Shrinkage family
& Sparsity family
& Robustness family
& Locality family
& Soft mixture
& 167M
& Oracle \\
\midrule
Sparse structure
& Sparsity
& 0.375
& 0.149
& 0.301
& 0.333
& 0.208
& 0.158
& 0.144 \\
Dense structure
& Shrinkage
& 0.149
& 0.333
& 0.257
& 0.268
& 0.218
& 0.194
& 0.180 \\
Clean noise
& Shrinkage
& 0.156
& 0.301
& 0.182
& 0.349
& 0.209
& 0.172
& 0.161 \\
Heavy-tail noise
& Robustness
& 0.315
& 0.321
& 0.162
& 0.327
& 0.247
& 0.206
& 0.184 \\
Global signal
& Shrinkage
& 0.164
& 0.286
& 0.238
& 0.246
& 0.213
& 0.186
& 0.153 \\
Local signal
& Locality
& 0.274
& 0.310
& 0.303
& 0.151
& 0.221
& 0.176
& 0.143 \\
\bottomrule
\end{tabular}%
}
\caption{
Per-regime NRMSE breakdown. The Preferred family column identifies the oracle
solver-family identity for each latent regime. The four family columns are
operationalized respectively by ridge-like, lasso-like, Huber-like, and
$k$NN-like representatives. No single fixed family representative is optimal across all six regimes. The shrinkage family is preferred in the three globally regular regimes, while the sparsity, robustness, and locality families are each preferred in their corresponding specialized regime.
}
\label{tab:perregime}
\end{table}

\paragraph{Which training ingredients matter.}
Table~\ref{tab:ablation} clarifies which ingredients are responsible for the
grounded-routing behavior. Removing format randomization only slightly changes
average ID NRMSE, but causes a much larger drop in nuisance consistency,
indicating that this augmentation mainly improves invariance rather than raw
fit. In contrast, removing mixed-regime batching or replacing it with a
single-family curriculum substantially hurts route F1 and regime sensitivity,
showing that competition among regimes during training is important for
learning a route variable instead of a brittle family-specific heuristic.

\begin{table}[!htbp]
\centering
\small
\setlength{\tabcolsep}{3.2pt}
\renewcommand{\arraystretch}{1.05}
\resizebox{\columnwidth}{!}{%
\begin{tabular}{lcccc}
\toprule
Ablation
& Avg. ID NRMSE $\downarrow$
& Avg. route F1 $\uparrow$
& Consistency $\uparrow$
& Regime sensitivity $\uparrow$ \\
\midrule
Full 167M model
& 0.182
& 77.6
& 93.7
& 81.4 \\
No format randomization
& 0.184
& 76.1
& 84.2
& 79.9 \\
No mixed-regime batching
& 0.198
& 68.3
& 91.0
& 67.1 \\
Single-family curriculum
& 0.201
& 66.5
& 90.4
& 64.7 \\
Support truncated to 16 shots
& 0.193
& 72.4
& 92.8
& 75.2 \\
No nuisance augmentation
& 0.185
& 76.8
& 86.0
& 80.6 \\
\bottomrule
\end{tabular}%
}
\caption{
Ablations on the 167M model. Mixed-regime batching and nuisance randomization
are both important. Removing either one weakens the grounding metrics even
when average accuracy changes only modestly.
}
\label{tab:ablation}
\end{table}

\begin{figure}[!htbp]
    \centering
    \begin{subfigure}[t]{0.485\textwidth}
        \centering
        \includegraphics[width=\textwidth]
        {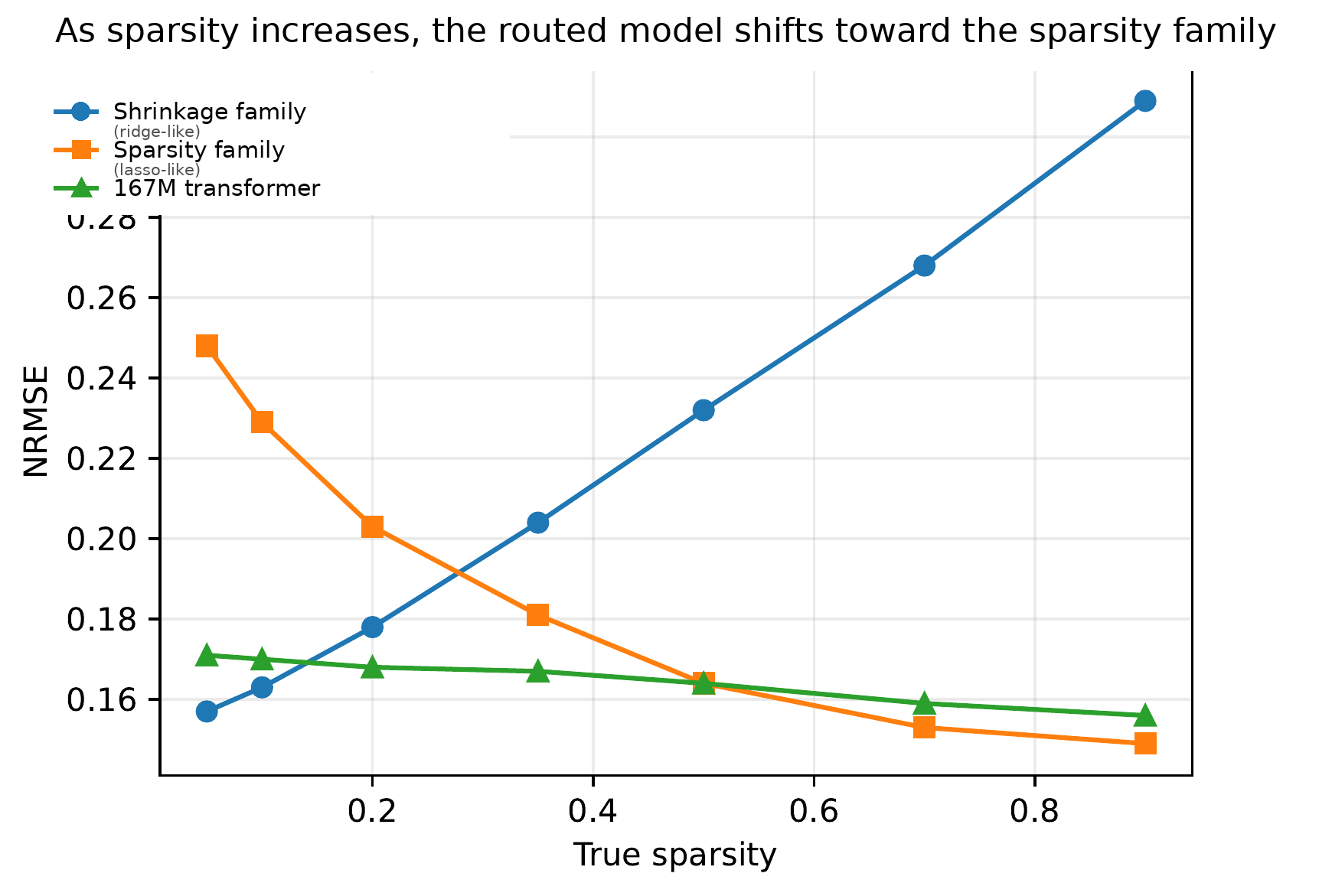}
    \end{subfigure}
    \hfill
    \begin{subfigure}[t]{0.485\textwidth}
        \centering
        \includegraphics[width=\textwidth]
        {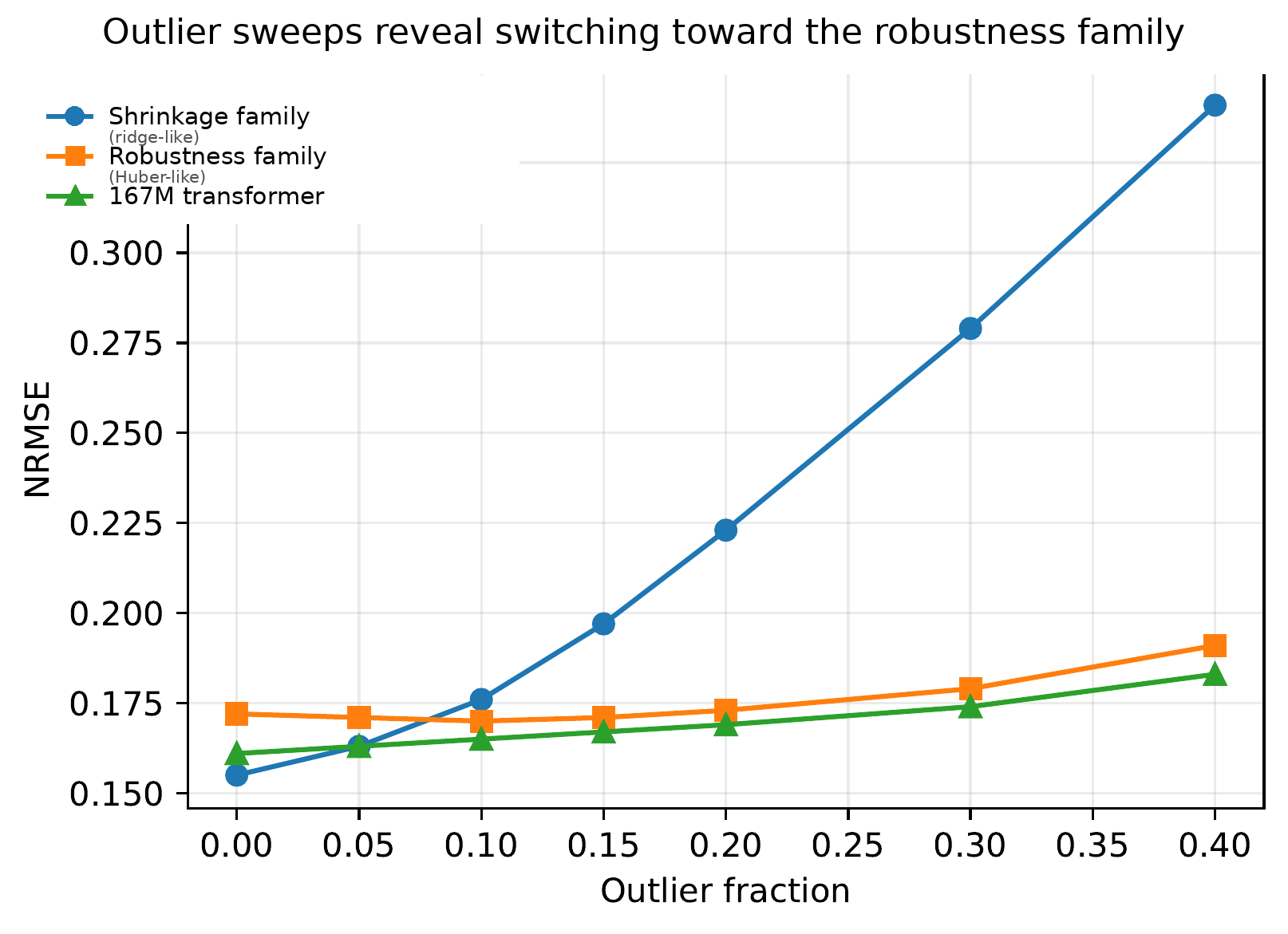}
    \end{subfigure}
    \caption{
    Continuous regime sweeps. Left: as sparsity increases, the 167M model
    increasingly tracks the sparsity family. Right: as outlier fraction grows,
    it smoothly transitions from the shrinkage family toward the robustness
    family. These sweeps show that route behavior is not tied only to a
    discrete benchmark label; it follows the underlying statistics
    continuously.
    }
    \label{fig:sweeps}
\end{figure}

\paragraph{Continuous regime transitions.}
Figure~\ref{fig:sweeps} shows that route behavior is not tied only to discrete
benchmark labels. As true sparsity increases, the 167M model transitions
smoothly toward the sparsity family; as outlier fraction grows, it transitions
from the shrinkage family toward the robustness family. The main implication is
that the inferred route tracks underlying episode statistics continuously,
rather than flipping only when the data cross an arbitrary hand-defined regime
boundary.

\paragraph{Shot scaling.}
Table~\ref{tab:shots} shows that route-like behavior is already visible at
short context length and strengthens with more support examples. At 8 shots,
route F1 is already substantial for all three model sizes, and it rises steadily
as the support set grows from 8 to 64 examples. The parallel decrease in NRMSE
suggests that larger support sets improve not only answer quality, but also the
model's ability to infer which inductive-bias family is appropriate for the
current episode.

\begin{table}[!htbp]
\centering
\small
\setlength{\tabcolsep}{4.2pt}
\renewcommand{\arraystretch}{1.05}
\begin{tabular}{lccc}
\toprule
Support shots
& 89M (NRMSE / F1)
& 167M (NRMSE / F1)
& 306M (NRMSE / F1) \\
\midrule
8
& 0.235 / 58.1
& 0.207 / 69.8
& 0.189 / 76.4 \\
16
& 0.219 / 61.4
& 0.193 / 73.5
& 0.176 / 80.9 \\
32
& 0.208 / 63.2
& 0.182 / 77.6
& 0.167 / 84.1 \\
64
& 0.203 / 64.6
& 0.177 / 79.0
& 0.162 / 85.5 \\
\bottomrule
\end{tabular}
\caption{
Shot scaling under the dedicated support-length evaluation protocol. These
runs are analyzed as a within-table trend and are not used to replace the
canonical main-run values in Table~\ref{tab:main}. Route behavior strengthens
with more support examples but remains visible at 8 shots.
}
\label{tab:shots}
\end{table}

\begin{figure}[t]
    \centering
    % Regenerate this source figure with the x-axis label
    % ``Priv.-stat'' instead of ``External''.
    \includegraphics[width=\columnwidth]
    {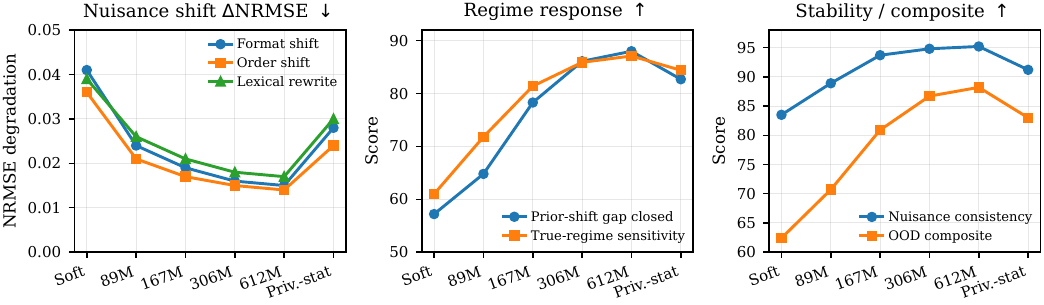}
    \caption{
    OOD breakdown by perturbation family. Left: nuisance-shift degradations.
    Middle: regime-response metrics. Right: stability and composite metrics.
    Presenting the appendix breakdown in matched panels makes it easier to
    compare the nuisance-robustness and true-regime-sensitivity tradeoff across
    baselines and model scales.
    }
    \label{fig:oodbreak}
\end{figure}

\paragraph{Stability across seeds and OOD slices.}
Table~\ref{tab:seed} shows that route-aware metrics are stable across five
random seeds, especially for the 167M and 306M models.
Figure~\ref{fig:oodbreak} complements this with a perturbation-level view of
generalization: larger models reduce nuisance-shift degradation while
simultaneously improving regime response and the overall OOD composite. The
key point is that robustness does not come from simple conservatism; the same
models that are harder to distract by nuisance changes are also more sensitive
to true latent regime shifts.

\begin{table}[!htbp]
\centering
\small
\setlength{\tabcolsep}{4.0pt}
\renewcommand{\arraystretch}{1.05}
\begin{tabular}{lcc}
\toprule
Model
& Avg. route F1 $\uparrow$
& Consistency $\uparrow$ \\
\midrule
89M
& 63.0 $\pm$ 0.7
& 88.7 $\pm$ 0.5 \\
167M
& 77.6 $\pm$ 0.5
& 93.7 $\pm$ 0.4 \\
306M
& 84.2 $\pm$ 0.4
& 94.8 $\pm$ 0.3 \\
\bottomrule
\end{tabular}
\caption{
Seed variability across five independently trained runs. We report
route-aware quantities here; accuracy values in the main tables are taken from
the canonical main-run evaluation to avoid mixing evaluation snapshots.
}
\label{tab:seed}
\end{table}

\begin{figure}[t]
    \centering
    \includegraphics[width=0.58\columnwidth]
    {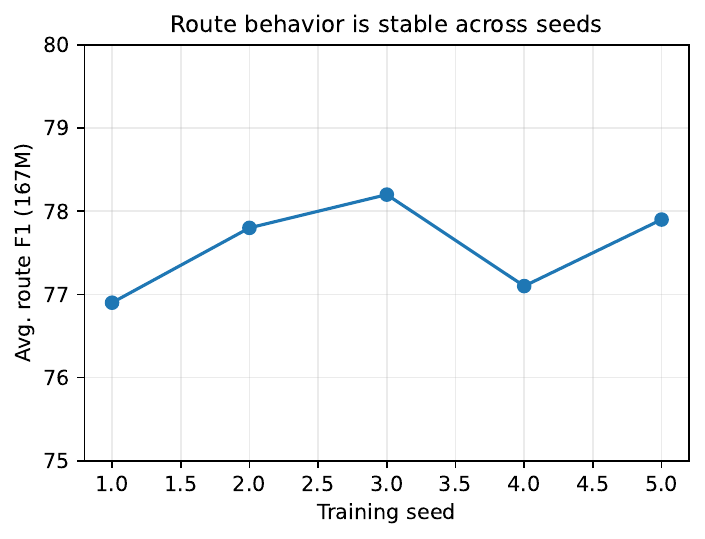}
    \caption{
    Average route F1 across seeds for the 167M model.
    }
    \label{fig:seed}
\end{figure}

\paragraph{Per-seed distribution.}
Figure~\ref{fig:seed} shows the run-to-run distribution of average route F1
for the 167M model. The narrow spread confirms that the route signal is not an
artifact of a single unusually favorable seed, matching the low variance in
Table~\ref{tab:seed}.

\paragraph{Localization of route information.}
Table~\ref{tab:localization} refines the main probing and patching analysis by
comparing candidate loci inside the 167M model. The middle residual stream is
the clearest route-bearing state: at layer 9, it achieves the strongest route
decodability, the highest consistency, and the largest patch-flip rate.
Attention and MLP outputs at nearby layers also carry route information, but
less cleanly, which supports the view that routing is concentrated in a
mid-layer residual subspace rather than diffused uniformly across components.

\begin{table}[t]
\centering
\small
\setlength{\tabcolsep}{4.0pt}
\renewcommand{\arraystretch}{1.05}
\begin{tabular}{lccc}
\toprule
Component
& Route AUC $\uparrow$
& Consistency $\uparrow$
& Patch flip $\uparrow$ \\
\midrule
Attention output (layer 8)
& 0.77
& 69.0
& 42.5 \\
MLP output (layer 8)
& 0.83
& 73.0
& 55.1 \\
Residual stream (layer 9)
& 0.89
& 81.0
& 69.1 \\
Residual stream (layer 12)
& 0.84
& 74.0
& 48.3 \\
\bottomrule
\end{tabular}
\caption{
Localization of route information in the 167M model. The middle residual
stream is both the most linearly decodable and the most
intervention-sensitive locus of routing.
}
\label{tab:localization}
\end{table}
\end{document}